\documentclass[conference]{IEEEtran}
\IEEEoverridecommandlockouts
\usepackage{cite}
\usepackage{amsmath,amssymb,amsfonts}
\usepackage{algorithm}
\usepackage{algorithmic}
\usepackage{graphicx}
\usepackage{textcomp}
\usepackage{xcolor}
\def\BibTeX{{\rm B\kern-.05em{\sc i\kern-.025em b}\kern-.08em
		T\kern-.1667em\lower.7ex\hbox{E}\kern-.125emX}}

\makeatletter
\let\old@ps@headings\ps@headings
\let\old@ps@IEEEtitlepagestyle\ps@IEEEtitlepagestyle
\def\confheader#1{%
	\def\ps@IEEEtitlepagestyle{
		\old@ps@IEEEtitlepagestyle
		\def\@oddhead{\strut\hfill#1\hfill\strut}
		\def\@evenhead{\strut\hfill#1\hfill\strut}
	}
	\ps@headings
}
\makeatother
\confheader{
	\small{Proceedings of the 13th RSI International Conference on Robotics and Mechatronics (ICRoM 2025), Dec. 16-18, 2025, Tehran, Iran} 
}
\usepackage[pscoord]{eso-pic}
\newcommand{\placetextbox}[3]{
	\setbox0=\hbox{#3}
	\AddToShipoutPictureFG*{ \put(\LenToUnit{#1\paperwidth},\LenToUnit{#2\paperheight}){\vtop{{\null}\makebox[0pt][c]{#3}}}
	}
}
\placetextbox{.5}{0.055}{\textbf{\small{Proceedings of the 13th RSI International Conference on Robotics and Mechatronics (ICRoM 2025), Dec. 16-18, 2025, Tehran, Iran}}}

\begin{document}

\title{Design of Adaptive PID Controller Based On Asynchronous Advantage Actor–Critic Learning Method for QuadCopter Control\\
}

\author{\IEEEauthorblockN{Ali Jokar}
\IEEEauthorblockA{\textit{Sharif AgRoLab} \\
\textit{School of Mechanical Engineering} \\
\textit{Sharif University of Technology}\\
Tehran, Iran \\
ali.jokar@sharif.edu}
\and
\IEEEauthorblockN{Aria Alasty}
\IEEEauthorblockA{\textit{Sharif AgRoLab} \\ 
\textit{School of Mechanical Engineering} \\
\textit{Sharif University of Technology}\\
Tehran, Iran \\
aalasti@sharif.edu}
}

\maketitle

\begin{abstract}
Quadcopters offer great utility in many applications, but their nonlinear nature and disturbance sensitivity present great control challenges. Basic PID controllers are generally not sophisticated enough to cope with these complexities. This paper suggests a control system that integrates the Asynchronous Advantage Actor-Critic (A3C) algorithm with a PID controller for quadcopter attitude and trajectory tracking. The A3C controller uses parallel agents to optimize PID parameters dynamically using a neural network. A system identification module for the complementary system makes predictions about system states for optimal control policy. The proposed framework was compared with a standard actor-critic (A2C) model. Simulation results verify that they both track accurately. However, the A3C-based controller converges much more for the loss function, as evidenced by reward figures and loss curves, demonstrating better parameter optimization. This shows that A3C-based approach results in improved performance for the control of quadcopter, effectively integrating reinforcement learning and traditional control to achieve higher adaptability.
\end{abstract}

\begin{IEEEkeywords}
QuadCopter, Deep Reinforcement Learning, A3C-PID, Adaptive PID Control
\end{IEEEkeywords}

\section{Introduction}
Quadcopters with their vertical take-off/landing capability and versatility have uses in agriculture, surveillance, and search and rescue operations but difficult to control due to them being nonlinear and disturbance-sensitive to wind and payload, for instance. Traditional PID controllers, although simple to design, are unable to deal with such issues \cite{b1}, \cite{b2}, \cite{b3}. Neural networks (NNs) and reinforcement learning (RL), actor-critic techniques in particular, surpass these limitations by capable of handling complicated functions and facilitate adaptive control through interaction with the environment \cite{b4}, \cite{b5}, \cite{b6}, \cite{b7}. Actor-critic techniques, thanks to their two-network configuration, assure stable training and effective exploration-exploitation for continuous control problems like quadcopter path tracking \cite{b8}, \cite{b9}, \cite{b10}.

Literature indicates RL-based adaptive PID controllers: Hernández-Alvarado et al. \cite{b4} employed NN-PID on underwater vehicles under disturbance robustness. De Paula et al. \cite{b5} utilized Q-learning for PID control of mobile robots for improved trajectory tracking. Sun et al. \cite{b8} utilized A3C-PID to stepping motors successfully in dealing with nonlinearity. Sharifi and Alasty \cite{b9} introduced a hybrid actor-critic NN architecture for quadcopters that provided self-tuning robustness. Adaptive RL-NN controllers combine RL for real-time adaptation and NNs for handling uncertainty to achieve greater robustness against nonlinearities and system changes \cite{b6}, \cite{b7}, \cite{b11}. Deep RL and DNNs improve control accuracy for stabilization and trajectory tracking \cite{b12}, \cite{b13}, \cite{b14}.

The paper proposes an improved A3C-PID scheme for quadcopter attitude control and trajectory tracking, combining A3C policy learning and PID robustness. Contributions include the combination of PID with deep RL, dynamic PID gain tuning via NNs, and minimizing computational complexity for real-time scenarios. The paper is organized in such a manner that Section II deals with quadcopter dynamics and PID control, Section III covers NN-based self-tuning PID, Section IV deals with the A3C-PID framework and optimization, Section V gives simulation results, and Section VI concludes with future work.

\section{Dynamic Modeling and PID Control}
\begin{figure}[htbp]
	\centerline{\includegraphics[width=7cm,height=4cm]{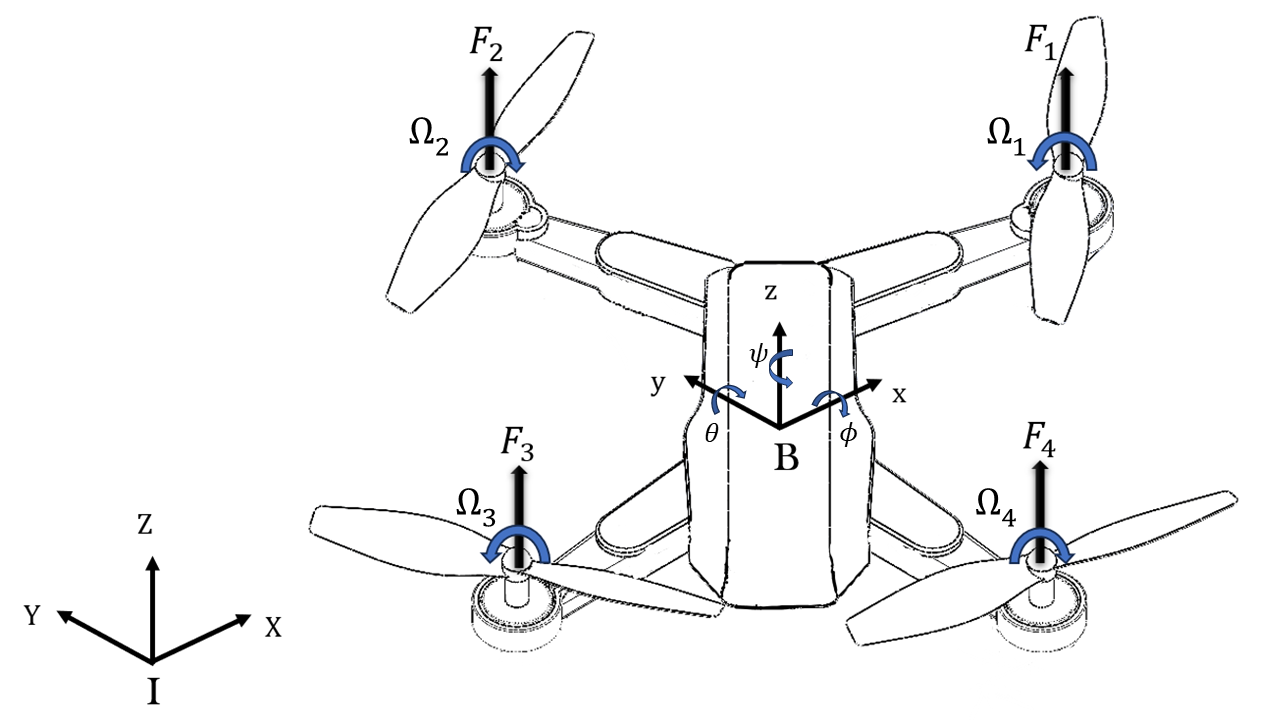}}
	\centering 
	\caption{Schematic of a quadcopter with coordinates}
	\label{quadcopter_schematic}
\end{figure}

Quadcopters such as one illustrated in Figure \ref{quadcopter_schematic}, are six degrees-of-freedom(DOF) under-actuated systems. Due to quadcopter inherent nonlinear dynamics and impacts from complicated environmental conditions, the system is hard to be modeled precisely. Therefore, applications of system identification methods, such as neural networks (NNs), have been a successful antidote in approximating the estimation of the system state.

A simplified mathematical model \cite{b9} is considered here in this research work to simulate the actual system free from noise.The equations of motion are given in Eq.1. In this formulation, ($x$, $y$, $z$) denote the positions of the center of gravity relative to the inertial reference coordinates ($x_I$, $y_I$, $z_I$), and ($\phi$, $\theta$, $\psi$) represent the rotational angles about the body axes ($x_B$,$y_B$,$z_B$):

\begin{equation}
\begin{aligned}
	\ddot{\phi} &= \frac{\dot{\theta}\dot{\psi}(J_y - J_z)}{J_x} + \frac{l}{J_x}u_2 ,\\
	\ddot{\theta} &= \frac{\dot{\phi}\dot{\psi}(J_z - J_x)}{J_y} + \frac{l}{J_y}u_3 ,\\
	\ddot{\psi} &= \frac{\dot{\phi}\dot{\theta}(J_x - J_y)}{J_z} + \frac{1}{J_z}u_4 ,\\
	\ddot{z} &= \frac{u_1}{m}\cos\phi\cos\theta - g, \\
	\ddot{x} &= \frac{u_1}{m}(\cos\phi\sin\theta\cos\psi + \sin\phi\sin\psi), \\
	\ddot{y} &= \frac{u_1}{m}(\cos\phi\sin\theta\sin\psi - \sin\phi\cos\psi),
\end{aligned}
\end{equation}
where $m$, $g$, and $l$ represent the total mass, gravitational acceleration, and arm length of the quadcopter, respectively. Additionally, $J_x$, $J_y$, and $J_z$ denote the moments of inertia around the principal axes of the body coordinate system.

The control inputs (Eq. 2) \(u_1\), \(u_2\), \(u_3\), and \(u_4\) are determined by the squared angular velocities of the motors (\(\Omega_1\), \(\Omega_2\), \(\Omega_3\), \(\Omega_4\)), with \(u_1\) affecting upward movement along the \(z\)-axis, \(u_2\) affecting roll movement, \(u_3\) affecting pitch movement, and \(u_4\) affecting yaw movement:

\begin{equation}
\begin{aligned}
	u_1 &= b(\Omega_1^2 + \Omega_2^2 + \Omega_3^2 + \Omega_4^2), \\
	u_2 &= b(\Omega_4^2 - \Omega_2^2), \\
	u_3 &= b(\Omega_3^2 - \Omega_1^2), \\
	u_4 &= d(\Omega_4^2 + \Omega_2^2 - \Omega_1^2 + \Omega_3^2),
\end{aligned}
\end{equation}
where $b$ and $d$ are the thrust and torque coefficients, respectively.

In this framework, the PID control algorithm is employed to generate the required control inputs in an online manner. Initially, static PID gains are selected using methods such as trial-and-error or the Ziegler-Nichols tuning method to achieve initial stability. These static gains are expressed as:

\begin{align}
	u_{\text{s}}(t) = K_p^s e(t) + K_i^s \int_0^t e(\tau)\,d\tau + K_d^s \dot{e}(t),
\end{align}
where $K_p^s$, $K_i^s$, and $K_d^s$ represent the static proportional, integral, and derivative gains, respectively.

To improve control performance, dynamically tuned gains are added to the static gains using a neural network based on the actor-critic method. The dynamic gains are expressed as:

\begin{align}
	u_{\text{d}}(t) = K_p^d e(t) + K_i^d \int_0^t e(\tau)\,d\tau + K_d^d \dot{e}(t),
\end{align}
where $K_p^d$, $K_i^d$, and $K_d^d$ are dynamically adjusted by the neural network. Thus, the total control input for each axis is given by:

\begin{align}
	u(t) = u_{\text{s}}(t) + u_{\text{d}}(t).
\end{align}

This process enables the controller to possess the ability to learn from system dynamics variation in a bid to increase stability and performance. Further details on the actor-critic process utilized for dynamic tuning will be outlined later in this paper.

\section{Network Structures}

The control scheme proposed in the present paper has two parts involve a self-tuning PID controller and a neural network-based system identification module. The two operate in concert to achieve adaptive and robust system performance.

\subsection{Self-Tuning PID Control}
The self-tuning PID controller uses a neural network for dynamic adaptive adjustment of proportional, integral, and derivative (PID) gains according to system states and error signals. The input to the network is the control inputs ($u(t-1)$, $u(t-2)$), system states ($s(t-1)$, $s(t-2)$), and PID error terms ($e_p(t-1)$, $e_i(t-1)$, $e_d(t-1)$). The output layer provides the adaptively adjusted PID gains ($K_p^d$, $K_i^d$, $K_d^d$), and they are determined as below:

\begin{align}
	\begin{split}
		K_p^d(t) = f_p(&u(t-1), u(t-2), s(t-1), s(t-2), \\
		&e_p(t-1), e_i(t-1), e_d(t-1)),
	\end{split}
\end{align}

\begin{align}
	\begin{split}
		K_i^d(t) = f_i(&u(t-1), u(t-2), s(t-1), s(t-2), \\
		&e_p(t-1), e_i(t-1), e_d(t-1)),
	\end{split}
\end{align}

\begin{align}
	\begin{split}
		K_d^d(t) = f_d(&u(t-1), u(t-2), s(t-1), s(t-2), \\
		&e_p(t-1), e_i(t-1), e_d(t-1)),
	\end{split}
\end{align}
where $f_p$,$f_i$ and $f_d$ represent a nonlinear mapping determined by the neural network. The activation functions in the hidden layers are sigmoid, while the output layer uses a $\tanh$ activation function to ensure smooth gain adjustments.

\begin{figure}[htbp]
	\centerline{\includegraphics[width=8cm,height=4.5cm]{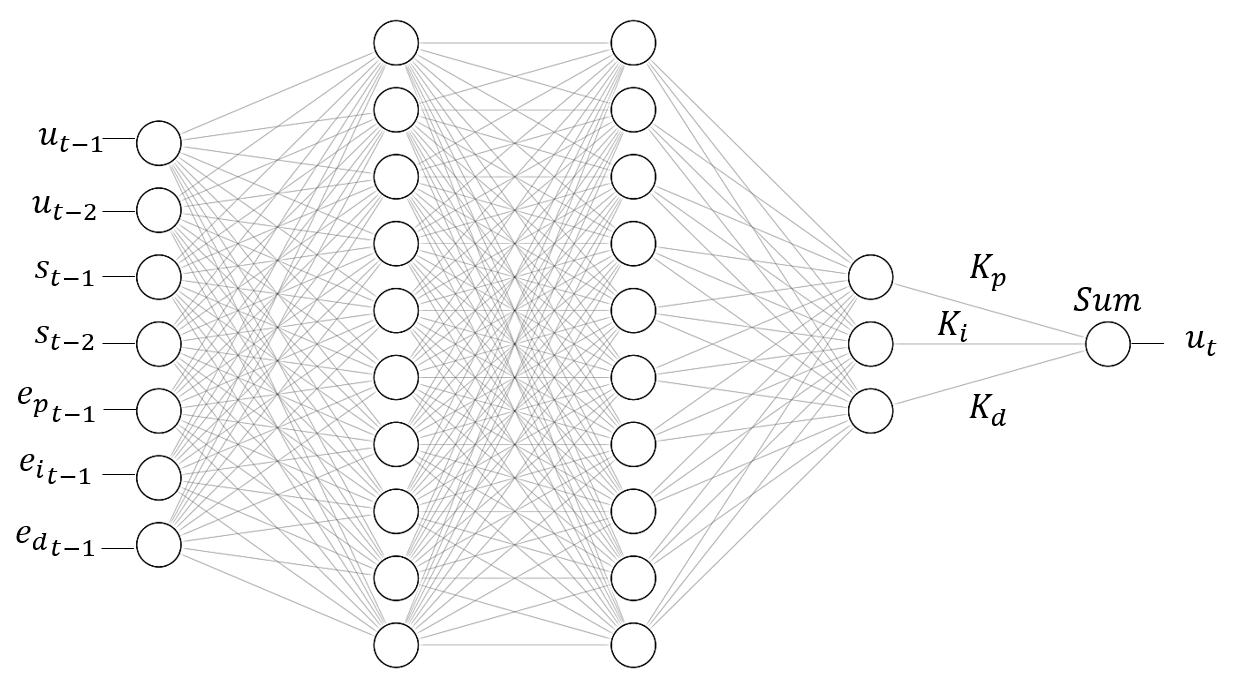}}
	\centering \caption{Self-tuning PID Neural Network Structure}
	\label{self1}
\end{figure}

\subsection{System Identification Using Neural Networks}
The system identification module employs an actor-critic structure to estimate the system's outputs. This module consists of two interconnected networks:

\begin{itemize}
	\item \textbf{Actor Network:} The actor network is responsible for estimating the system's state outputs. It has two outputs: an average ($\mu$) and a variance ($\sigma$), which define a Gaussian distribution:
	\begin{align}
		\mu(t) &= f_\mu(u(t), s(t-1), s(t-2)), \\
		\sigma(t) &= f_\sigma(u(t), s(t-1), s(t-2)),
	\end{align}
	A random sample drawn from this distribution represents the system's estimated output. The hidden layers use sigmoid activation functions.
	
	\begin{figure}[htbp]
		\centerline{\includegraphics[width=8cm,height=4cm]{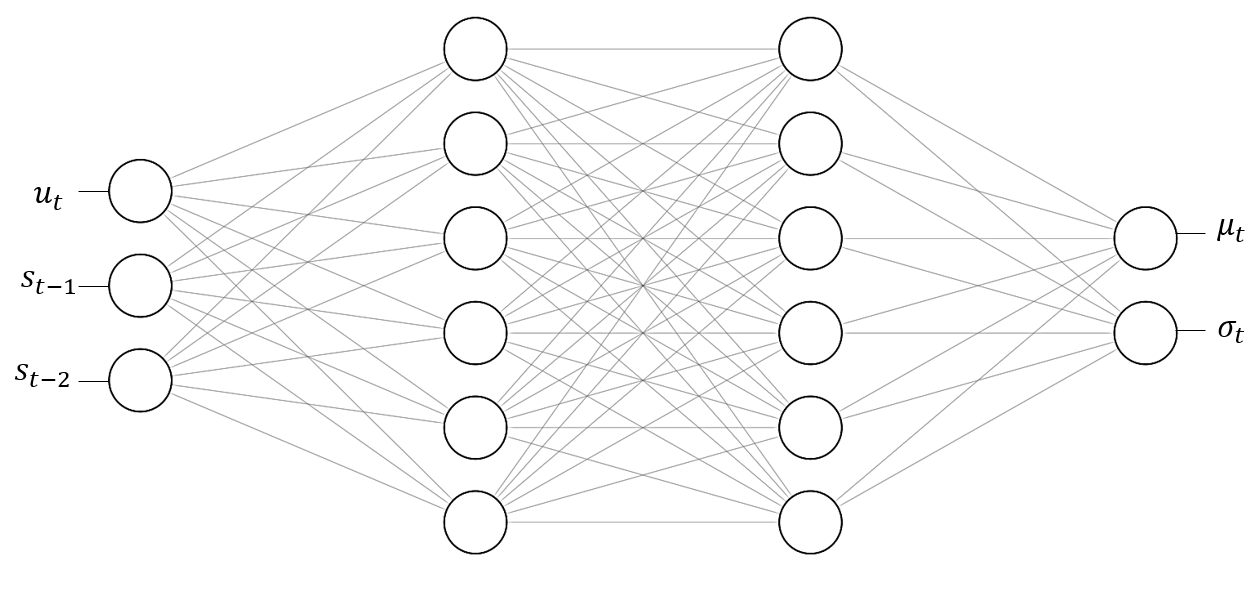}}
		\centering \caption{Actor Neural Network Structure}
		\label{self2}
	\end{figure}
	
	\item \textbf{Critic Network:} The critic network evaluates the value function ($v$) of the given state and control input. This value function provides feedback to the actor network, enabling it to improve its estimations:
	\begin{align}
		v(t) = f_v(u(t), s(t-1), s(t-2)),
	\end{align}
	Similar to the actor network, the hidden layers of the critic network also employ sigmoid activation functions.
	
	\begin{figure}[htbp]
		\centerline{\includegraphics[width=8cm,height=4.5cm]{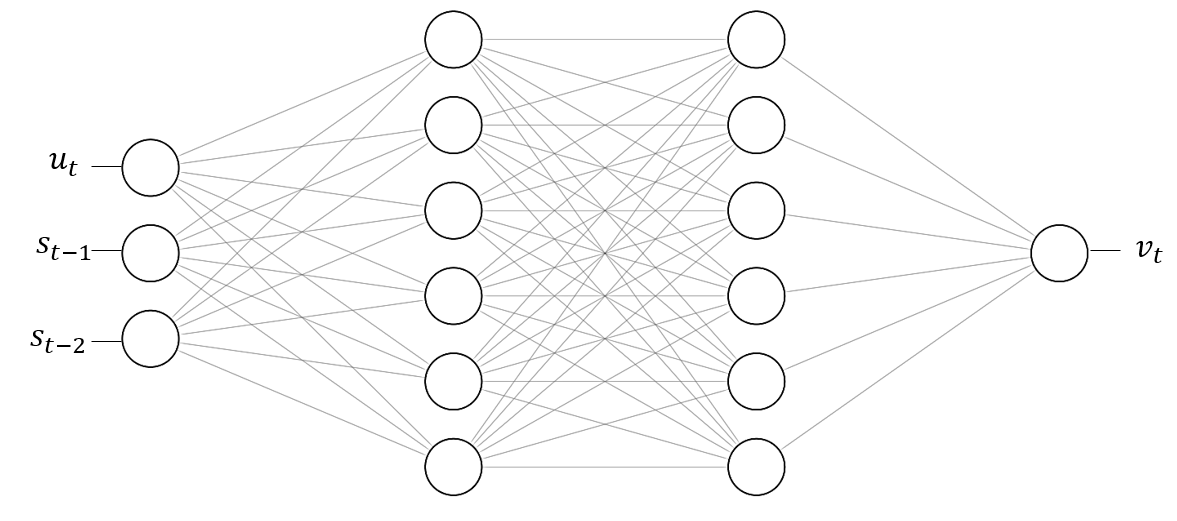}}
		\centering \caption{Critic Neural Network Structure}
		\label{self3}
	\end{figure}
\end{itemize}

\subsection{Integration of Networks}
In order to achieve an integrated control strategy with both, the system identification network and self-tuning network are series connected. The time-varying gains of the self-tuning network are used in computing the control inputs, and these control inputs are fed to the system identification network. The integration creates real-time flexibility and does not need a precise model of the system, making the approach model-free.

The total system identification network outputs are the state estimates and the value function of the input states. The two provide the necessary flexibility to control nonlinear and dynamic systems such as quadcopters.

\section{A3C Algorithm}
\subsection{Reward and Optimization formulas}
Following architecture and network structure determination, network parameter weights and bias would still need to be further optimized using an optimization algorithm. The actor network would attempt to reduce estimation error between actual state ($s$) and model-estimated state ($s_m$), and the critic network would attempt to reduce Temporal Difference (TD) error ($\delta_{TD}$). TD error can be defined as:

\begin{align}
	\delta_{TD} = R_{t+1} + \gamma v_{t+1} - v_t,
\end{align}
where $\gamma$ is a discount factor, $R_{t+1}$ is a reward function, and $v_t$ and $v_{t+1}$ are value functions at time $t$ and time $t+1$, respectively.

The reward function promotes minimum absolute error, low error rates and minimum control input. It can be defined as:

\begin{align}
	R_{t+1} = -\kappa_1(s_m - s)^2 - \kappa_2(\dot{s}_m - \dot{s})^2 - \kappa_3 u^2,
\end{align}
where $\kappa_1$, $\kappa_2$, and $\kappa_3$ are weight coefficients, and $u$ is the control signal. These terms ensure that the network balances precision, stability, and energy efficiency.

Separate loss functions are defined for the actor ($L_a$) and the critic ($L_c$):

\begin{align}
	L_a &= \lambda_1 (s_m - s)^2 ({\zeta + |\delta_{TD}|}) + \lambda_2 \sqrt{2\pi e^{\sigma^2}}, \\
	L_c &= \lambda_3 \delta_{TD}^2,
\end{align}
where $\lambda_1$, $\lambda_2$, and $\lambda_3$ are the weights, and $zeta$ is a small value to prevent division by zero. The TD error is calculated by the critic network and is backpropagated to the actor network. The greater the TD error, the greater the actor loss, in the expectation of encouraging the actor to revise its action. When TD error is zero, that is, it has reached optimal actions.

Total loss function is a combination of actor and critic losses:

\begin{align}
	L_{total} = L_a + L_c.
\end{align}

For the best network parameters, ADAM optimizer \cite{b15} is employed because it is both effective and stable in handling deep neural networks. The update rules of ADAM are provided as:

\begin{align}
	m_t &= \beta_1 m_{t-1} + (1 - \beta_1) g_t, \\
	v_t &= \beta_2 v_{t-1} + (1 - \beta_2) g_t^2, \\
	\hat{m}_t &= \frac{m_t}{1 - \beta_1^t}, \quad \hat{v}_t = \frac{v_t}{1 - \beta_2^t}, \\
	\theta_{t+1} &= \theta_t - \frac{\alpha}{\sqrt{\hat{v}_t} + \epsilon} \hat{m}_t,
\end{align}
where $g_t = \frac{\partial L_a}{\partial \theta_t}$ is the gradient of the loss function with respect to the parameters, $\alpha$ is the learning rate, and $\epsilon$ is a small constant for numerical stability.

The gradients for the self-tuning ($\theta_{st}$) and system identification ($\theta_{si}$) networks are obtained as:

\begin{align}
	g_{st} &= \left(\frac{\partial L_a}{\partial s_m} \frac{\partial s_m}{\partial u} + \frac{\partial L_a}{\partial \sigma} \frac{\partial \sigma}{\partial u} + \frac{\partial L_c}{\partial v} \frac{\partial v}{\partial u}\right) \frac{\partial u}{\partial \theta_{st}}, \\
	g_{si} &= \frac{\partial L_a}{\partial s_m} \frac{\partial s_m}{\partial \theta_{si}} + \frac{\partial L_a}{\partial \sigma} \frac{\partial \sigma}{\partial \theta_{si}} + \frac{\partial L_c}{\partial v} \frac{\partial v}{\partial \theta_{si}}.
\end{align}

The partial derivative $\frac{\partial u}{\partial \theta_{st}}$ is computed as:

\begin{align}
	\frac{\partial u}{\partial \theta_{st}} = e_p \frac{\partial K_p^d}{\partial \theta_{st}} + e_i \frac{\partial K_i^d}{\partial \theta_{st}} + e_d \frac{\partial K_d^d}{\partial \theta_{st}},
\end{align}
where $e_p$, $e_i$, and $e_d$ are externally injected errors and non-trainable. The proposed structure, although designed specifically for Single-Input-Single-Output (SISO) systems, can be extended for Multi-Input-Multi-Output (MIMO) systems such as quadcopters by decoupling the dynamics into separate subsystems for every state variable (i.e., $\phi$, $\theta$, $\psi$, $z$).

\subsection{A3C Framework}
Asynchronous advantage actor-critic (A3C) framework enhances solution flexibility offered. A3C exploits multiple parallel environments to reduce sample correlation during training and improve convergence speed. Unlike its synchronous counterpart A2C (Advantage Actor-Critic) with a single agent, A3C employs a number of concurrent workers. The actor-critic algorithm in the two models has two networks: an actor network that produces the optimal actions for the given state, and a critic network that provides an estimation of the value function for state-action pairs. 

A3C's asynchronous implementation ensures computational effectiveness, particularly in real-time control systems, while A2C ensures simpler implementation with sequential sampling. The algorithm, as that of Algorithm 1, explains how multiple workers of A3C update global network weights asynchronously using parallel sampling and gradient computation, while A2C update would be synchronous. With the integration of A3C in self-tuning PID and system identification networks, the proposed approach is achieved with improved stability and performance under various conditions.

\begin{algorithm}
	\caption{A3C (Asynchronous Advantage Actor-Critic)\cite{b16}}
	\begin{algorithmic}[1]
		\STATE \textbf{Master:}
		\STATE \textbf{Hyperparameters:} Step sizes $\eta_\psi$ and $\eta_\theta$, current policy $\pi_\theta$, value function $V^{\pi_\theta}_\psi$.
		\STATE \textbf{Input:} Gradients $g_\psi$, $g_\theta$.
		\STATE Update parameters:
		\begin{align*}
			\psi &\gets \psi - \eta_\psi g_\psi, \\
			\theta &\gets \theta + \eta_\theta g_\theta.
		\end{align*}
		\STATE Return $(V^{\pi_\theta}_\psi, \pi_\theta)$.
		
		\STATE
		\STATE \textbf{Worker:}
		\STATE \textbf{Hyperparameters:} Reward discount factor $\gamma$, trajectory length $L$.
		\STATE \textbf{Input:} Value function $V^{\pi_\theta}_\psi$, policy $\pi_\theta$.
		\STATE Initialize gradients: $(g_\theta, g_\psi) \gets (0, 0)$.
		\FOR{$k = 1, 2, \ldots$}
		\STATE Synchronize parameters: $(\theta, \psi) \gets \text{Master}(g_\theta, g_\psi)$.
		\STATE Run policy $\pi_\theta$ for $L$ time steps and collect transitions $\{S_t, A_t, R_t, S_{t+1}\}$.
		\STATE Estimate advantages:
		\begin{align*}
			\hat{A}_t = R_t + \gamma V^{\pi_\theta}_\psi(S_{t+1}) - V^{\pi_\theta}_\psi(S_t).
		\end{align*}
		\STATE Compute objective functions:
		\begin{align*}
			J(\theta) &= \sum_t \log \pi_\theta(A_t | S_t) \hat{A}_t, \\
			J_{V^{\pi_\theta}_\psi}(\psi) &= \sum_t \hat{A}_t^2.
		\end{align*}
		\STATE Compute gradients:
		\begin{align*}
			g_\psi &\gets \nabla J_{V^{\pi_\theta}_\psi}(\psi), \\
			g_\theta &\gets \nabla J(\theta).
		\end{align*}
		\ENDFOR
	\end{algorithmic}
\end{algorithm}

\section{Results and Discussion}
\begin{figure}[H]
	\centerline{\includegraphics[width=9cm,height=4.1cm]{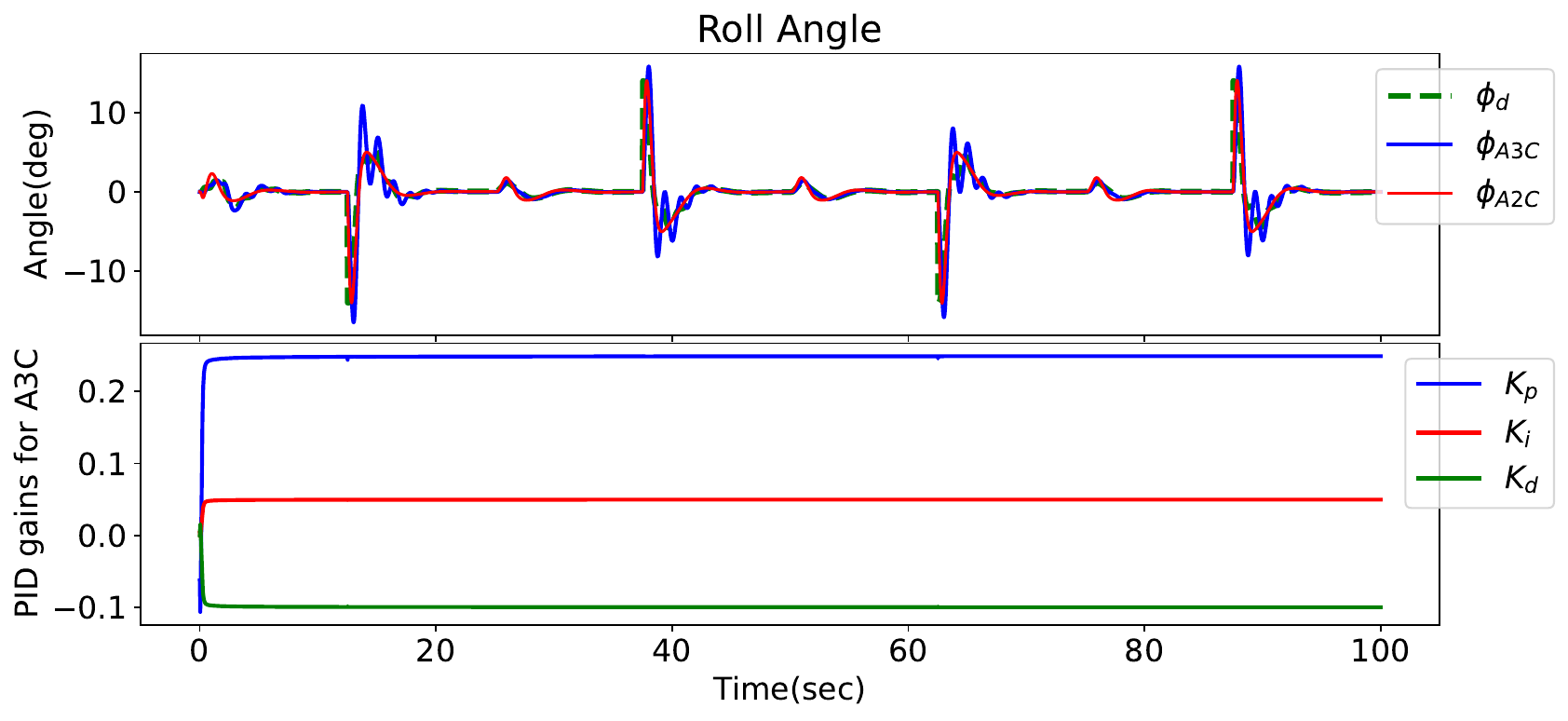}}
	\centering \caption{Comparison of Roll ($\phi$) Angle Control Using A2C and A3C Algorithms}
	\label{4_1}
\end{figure}

\begin{figure}[H]
	\centerline{\includegraphics[width=9cm,height=4.1cm]{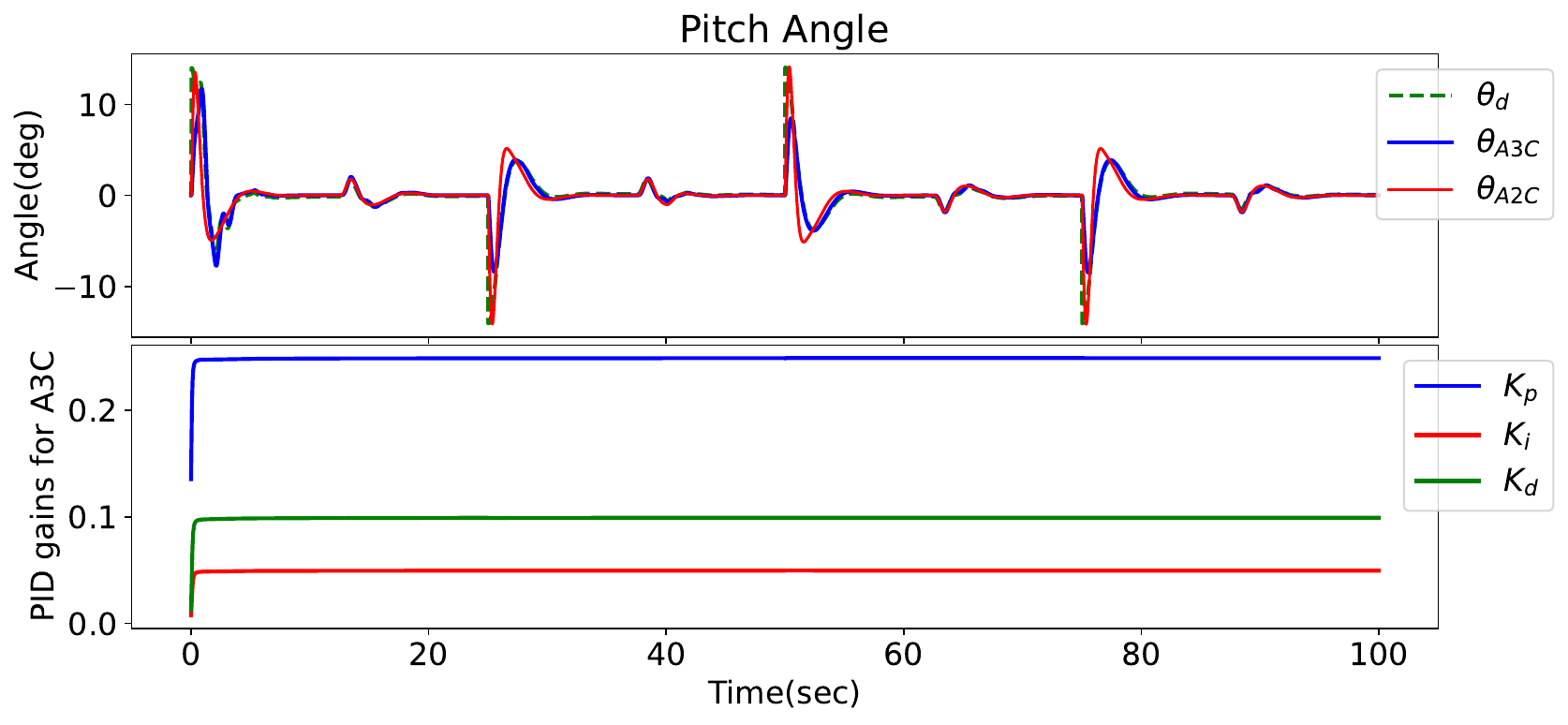}}
	\centering \caption{Comparison of pitch ($\theta$) Angle Control Using A2C and A3C Algorithms}
	\label{5_1}
\end{figure}

\begin{figure}[H]
	\centerline{\includegraphics[width=9cm,height=4.1cm]{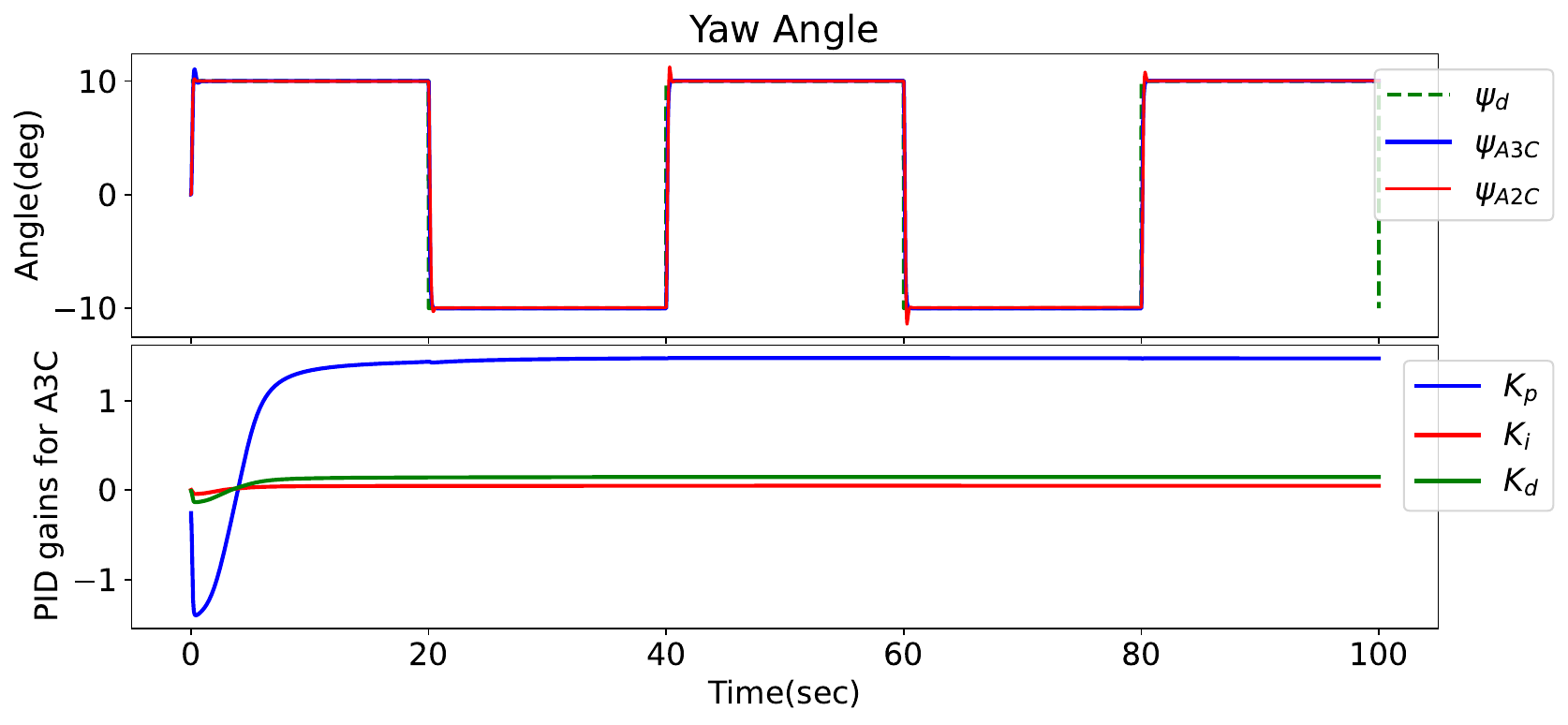}}
	\centering \caption{Comparison of yaw ($\psi$) Angle Control Using A2C and A3C Algorithms}
	\label{6_1}
\end{figure}

Fixed height squared trajectory was selected for this experiment to evaluate the proposed control strategy through the disturbance of quadcopter stability and path-tracking ability. The system began with initial gains but were soon tuned by the agent to optimal values, indicating good parameter tuning. With accumulation of experience, the adjustments became stabilized, which confirmed convergence to optimal gains and good adaptation towards system dynamics [Figs.~\ref{4_1}, \ref{5_1}, \ref{6_1}].

\begin{figure}[H]
	\centerline{\includegraphics[width=9cm,height=4cm]{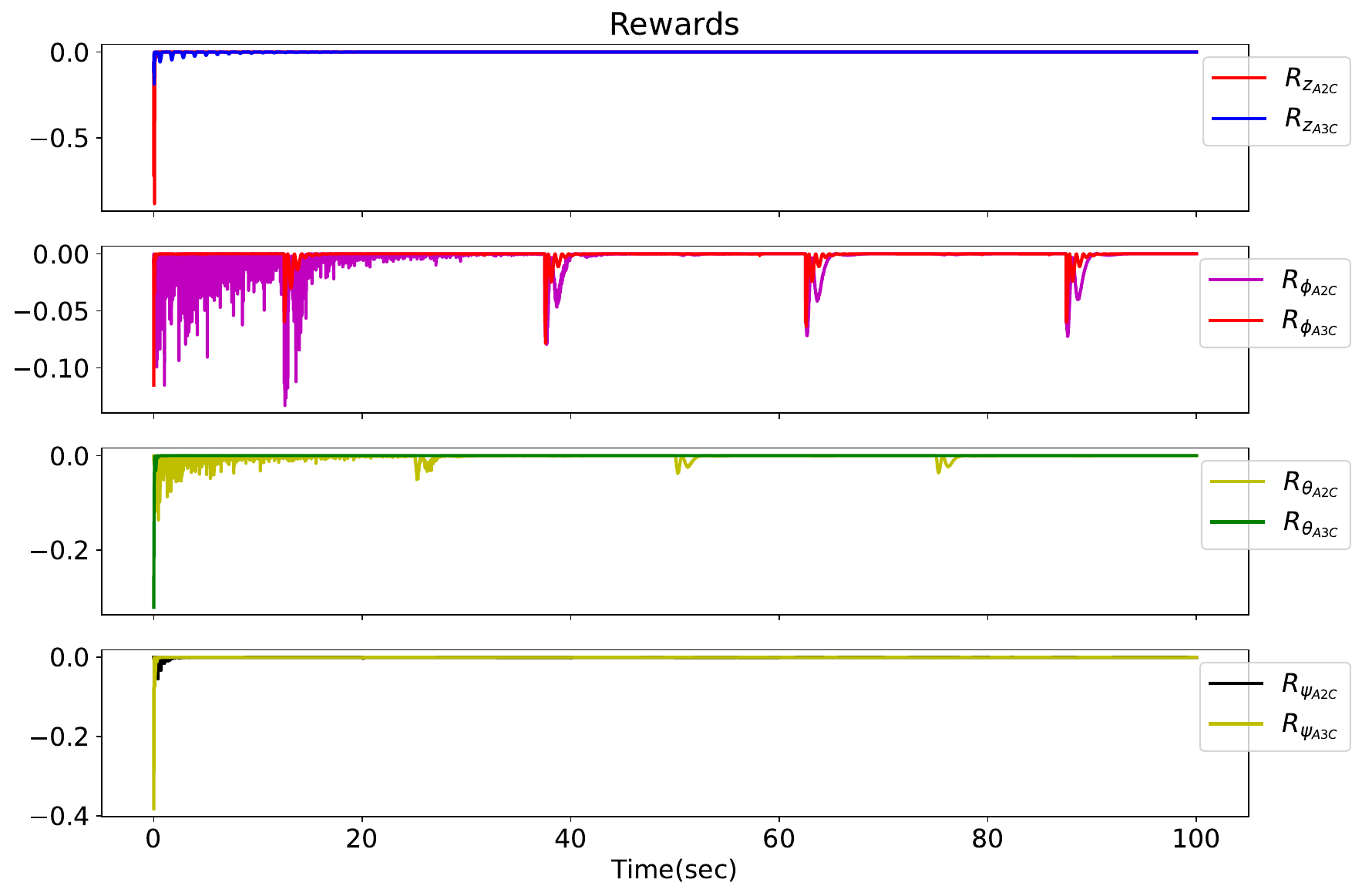}}
	\centering \caption{Comparison of Received rewards of each agent Using A2C and A3C Algorithms}
	\label{7_1}
\end{figure}

\begin{figure}[H]
	\centerline{\includegraphics[width=9cm,height=4.2cm]{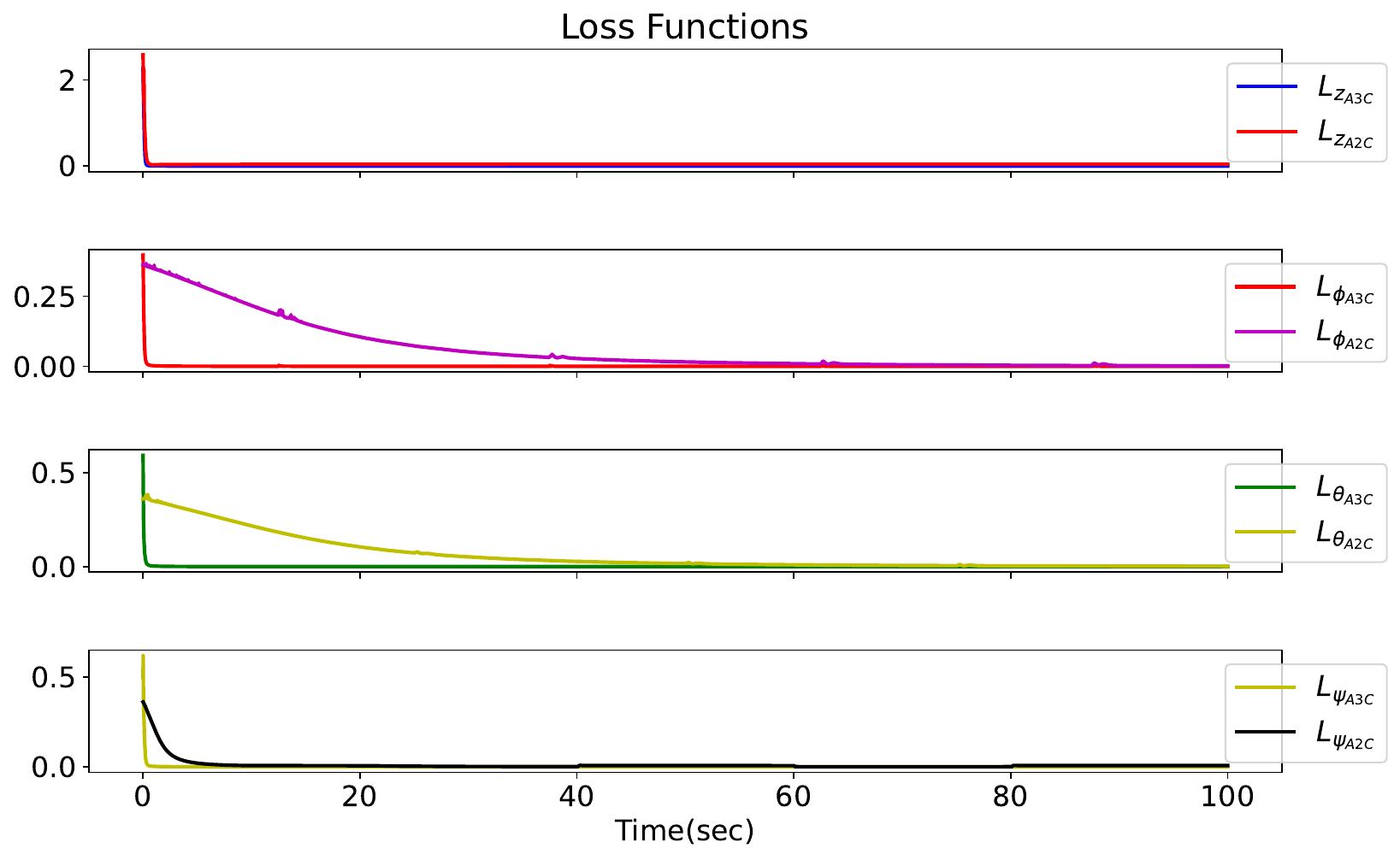}}
	\centering \caption{Comparison of Loss Function Variations in Each Agent’s Network Using A2C and A3C Algorithms}
	\label{8_1}
\end{figure}

Improving reward values and reducing loss functions validated the optimization process, illustrating real-time network parameter tuning and rapid convergence to optimal control. This efficiency is suitable for dynamic robotic systems [Figs.~\ref{7_1}, \ref{8_1}].

\begin{figure}[H]
	\centerline{\includegraphics[width=8cm,height=4.2cm]{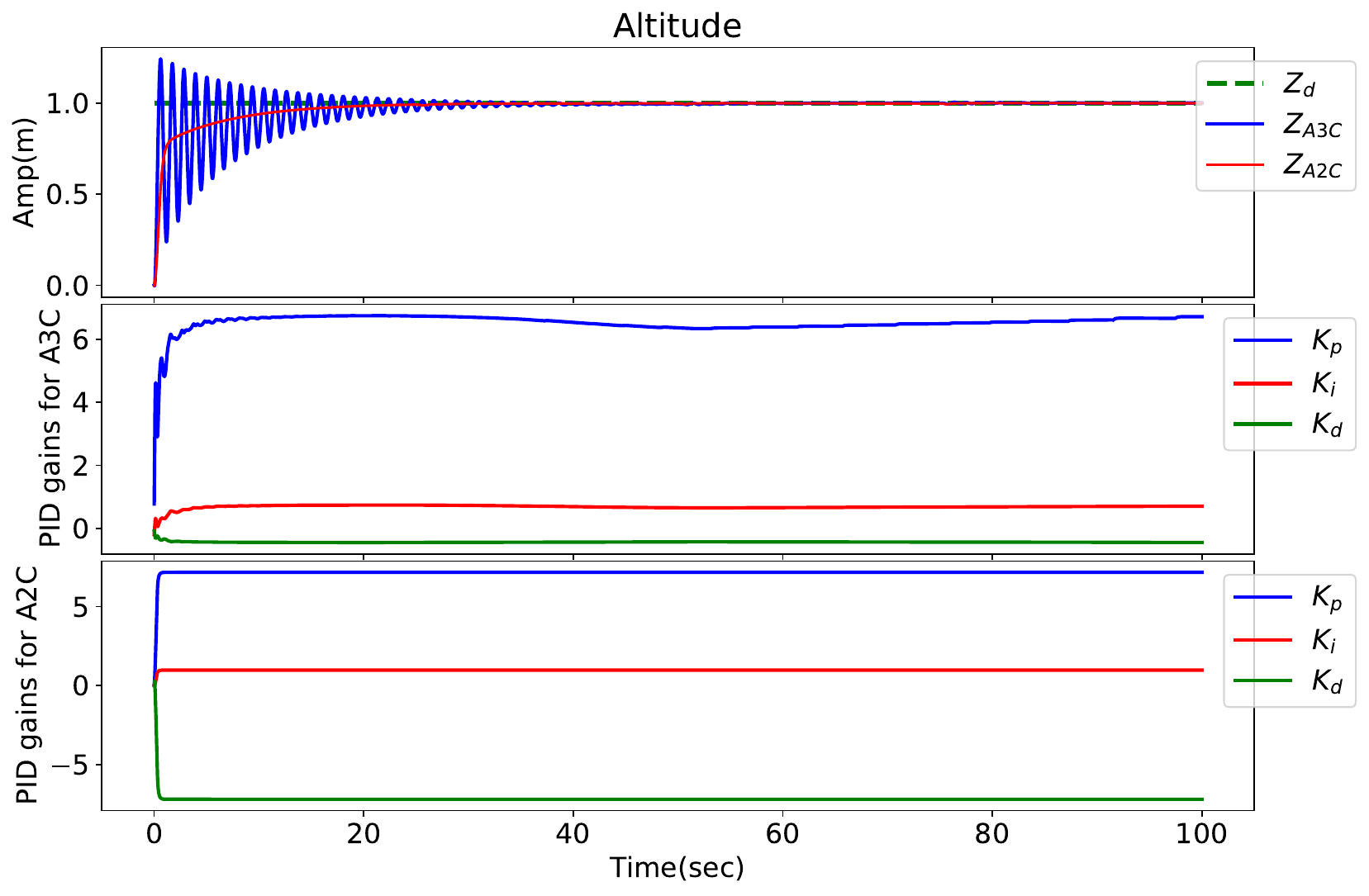}}
	\centering \caption{Comparison of tracking Desired Z path for each A2C and A3C Algorithms}
	\label{3_1}
\end{figure}

\begin{figure}[H]
	\centerline{\includegraphics[width=8cm,height=4.2cm]{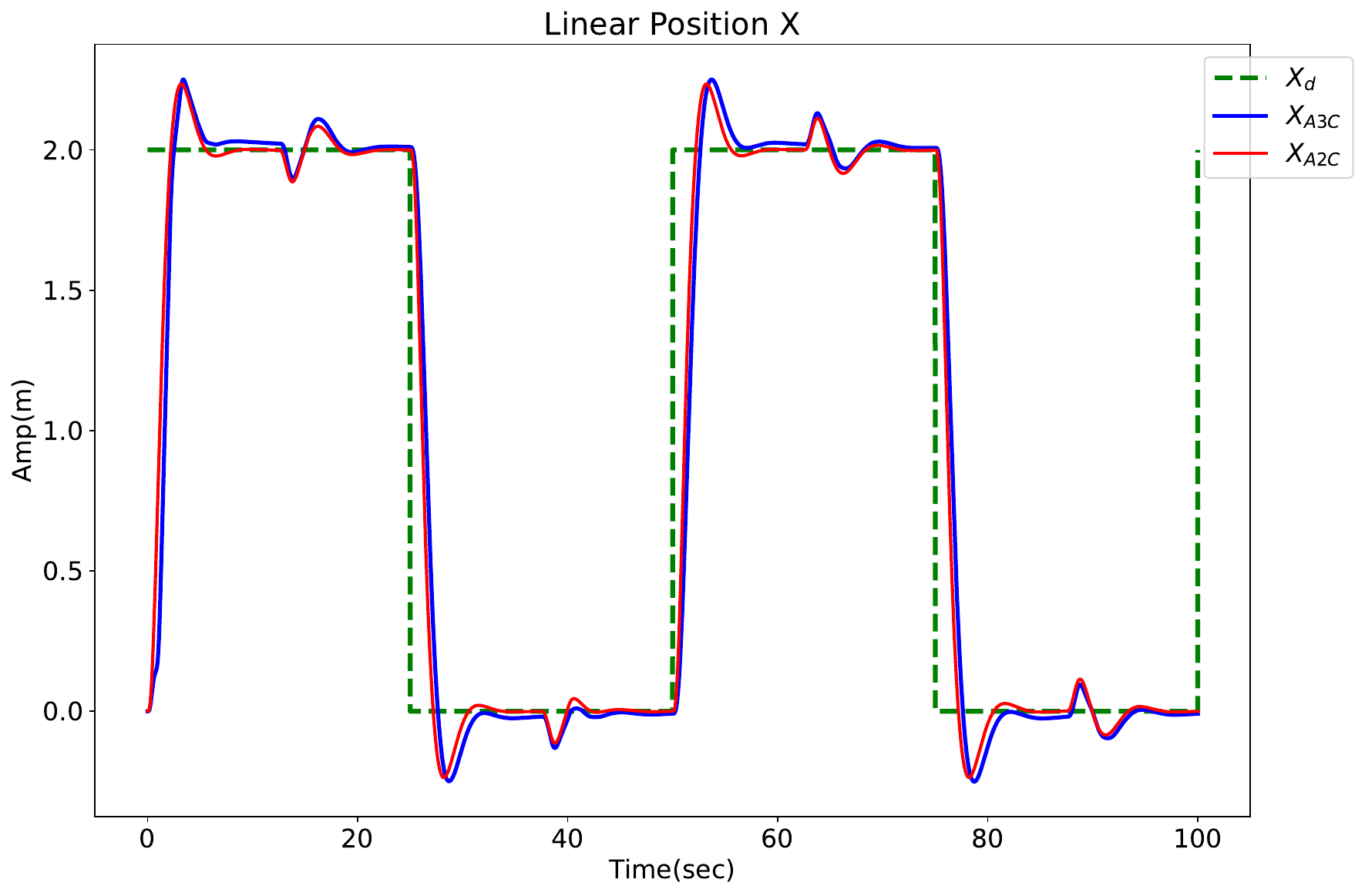}}
	\centering \caption{Comparison of tracking Desired X path for each A2C and A3C Algorithms}
	\label{1_1}
\end{figure}

\begin{figure}[H]
	\centerline{\includegraphics[width=8cm,height=4.2cm]{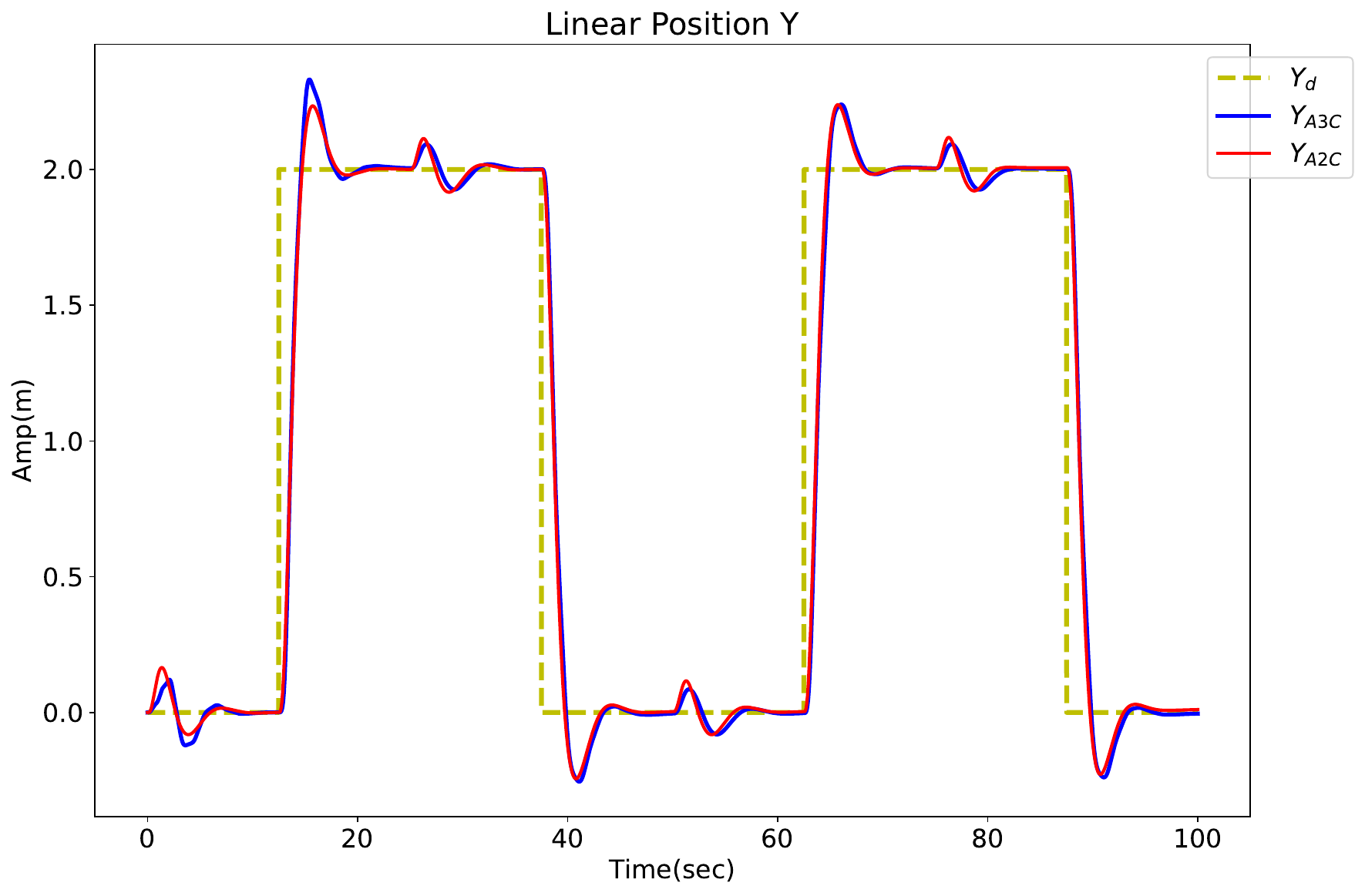}}
	\centering \caption{Comparison of tracking Desired Y path for each A2C and A3C Algorithms}
	\label{2_1}
\end{figure}

Comparison of A2C and A3C for PID gain tuning:
\begin{itemize}
	\item \textbf{Altitude (z-axis)}: A3C oscillated more at the beginning before converging, while A2C approached steadily. While A3C tracked closer, its overshoot may be undesirable for stability-critical applications [Fig.~\ref{3_1}].
	\item \textbf{Position tracking (x/y-axes)}: A2C oscillated less, yet they showed almost the same performance [Figs.~\ref{1_1}, \ref{2_1}].
	\item \textbf{Roll/pitch/yaw angles}: A3C experienced greater oscillations, while A2C experienced smoother dynamics with less fluctuation [Figs.~\ref{4_1}, \ref{5_1}, \ref{6_1}].
	\item \textbf{Reward trends}: A2C had higher initial variability, but A3C converged faster to optimal rewards [Fig.~\ref{7_1}].
	\item \textbf{Loss functions}: A3C dropped faster than A2C, confirming faster optimization [Fig.~\ref{8_1}].
\end{itemize}

It is in this aspect that the system identification module acts as a vital building block, providing the A3C controller with predictive state estimates that enhance its robustness against nonlinear quadcopter dynamics and external disturbances. In comparison with A2C, the A3C model has faster convergence of the loss function because of its asynchronous architecture; the optimization process speeds up due to lesser sample correlation since there are many agents working in parallel. This advantage comes at a cost, though: while A3C achieves higher tracking accuracy eventually because of this faster convergence, larger initial oscillations characterize A3C as compared to the smoother performance of the A2C controller. This signifies that the choice between A3C and A2C should be application-specific, trading between high precision and smooth control.

The method possesses great flexibility, tracking accuracy, and convergence speed but is plagued by altitude oscillations, computations, and dependency on data. Future work will focus on improving altitude stability, real-world applicability, algorithm fusion, and power efficiency. 

Generally, A2C and A3C both show great performance on PID tuning; A3C converges faster and has more precise tracking, which can be employed in precision-critical missions, while A2C realizes smoother control with minimal overshoot, which is suitable for stability-critical missions.

\section{Conclusion}
The article suggested an adaptive PID controller for quadcopters using A3C to overcome the limitations of traditional PID controllers in dealing with nonlinear dynamics and disturbances. The A3C-PID controller was trained to update gains dynamically using deep reinforcement learning, while its actor-critic module facilitated state estimation without relying on realistic models.

Simulations indicated that A3C-PID outperformed A2C in trajectory tracking and attitude tracking by having a faster convergence as well as higher accuracy. But it caused larger oscillations, particularly holding altitude, than A2C, but with smoother performance. It was a trade-off between smoothness and convergence rate, and application-specific tuning needed to be performed.

As robust as it was, A3C-PID's oscillation had to be enhanced by reward function and tuning of optimization. Future work needs to concentrate on investigating hybrid A3C-A2C methods, increasing the MIMO system, evaluating under noisy and disrupted conditions of the real world, uniting energy-efficient control and obstacle avoidance, and further utilization in search and rescue operations, surveillance, and delivery.

The A3C-PID controller represented a significant advancement from adaptive quadcopter control in declining conditions. Although there still existed potential for optimization, the study demonstrated the effectiveness of using reinforcement learning along with traditional control methods. The future studies were poised to address performance improvement and real-world capability enhancement.

\vspace{12pt}

\end{document}